\documentclass[preprints,article,accept,pdftex,moreauthors]{Definitions/mdpi} 
\usepackage{subcaption}

\firstpage{1} 
\pubvolume{1}
\issuenum{1}
\articlenumber{0}
\pubyear{2024}
\copyrightyear{2024}
\datereceived{ } 
\daterevised{ } 
\dateaccepted{ } 
\datepublished{ } 
\hreflink{https://doi.org/} 

\Title{Synthesizer Based Efficient Self-Attention for Vision Tasks}
\TitleCitation{Synthesizer Based Efficient Self-Attention for Vision Tasks}

\Author{Guangyang Zhu $^{1}$\orcidA{}, Jianfeng Zhang $^{2}$*\orcidB{}, Yuanzhi Feng $^{3}$\orcidC{} and Hai Lan $^{4}$}

\AuthorNames{Guangyang Zhu, Jianfeng Zhang, Yuanzhi Feng and Hai Lan}

\AuthorCitation{Zhu, G.; Zhang, J.; Feng, Y.; Lan, H.}

\address{%
$^{1}$ \quad School of Geospatial Information, Information Engineering University, Zhengzhou 450052, China; zhuguangyang13@126.com\\
$^{2}$ \quad Department of Information, East China Normal University, Shanghai 200241, China; zzhangjf1995@gmail.com\\
$^{3}$ \quad Software Engineering Institute, East China Normal University, Shanghai 200241, China; fengyuanzhi21@mails.ucas.ac.cn\\
$^{4}$ \quad Fujian Institute of Research on the Structure of Matter, Chinese Academy of Sciences, Quanzhou 362200, China; lanhai09@fjirsm.ac.cn
}

\corres{Correspondence: zzhangjf1995@gmail.com}

\abstract{Self-attention module shows outstanding competence in capturing long-range relationships while enhancing performance on vision tasks, such as image classification and image captioning. 
However, the self-attention module highly relies on the dot product multiplication and dimension alignment among query-key-value features, which cause two problems: (1) The dot product multiplication results in exhaustive and redundant computation.
(2) Due to the visual feature map often appearing as a multi-dimensional tensor, reshaping the scale of the tensor feature to adapt to the dimension alignment might destroy the internal structure of the tensor feature map. 
To address these problems, this paper proposes a self-attention plug-in module with its variants,
namely, Synthesizing Tensor Transformations (STT), for directly processing image tensor features. 
Without computing the dot-product multiplication among query-key-value, the basic STT is composed of the tensor transformation to learn the synthetic attention weight from visual information. 
The effectiveness of STT series is validated on the image classification and image caption.
Experiments show that the proposed STT achieves competitive performance while keeping robustness compared to self-attention in the aforementioned vision tasks.}


\begin{document}


\section{Introduction}

In recent years, the attention mechanism is generating considerable interest in the domain of deep learning. The main breakthroughs of attention modules were first appeared in Natural Language Processing (NLP) \cite{routeformer,bigbird,reformer,memformer,performer,compressive}, and then also in computer vision domain \cite{detectdeformable,huang2019attention_AOA_iccv,FAHIM2020106437,tong2024edge}. These achievements have demonstrated that the attention module provided a different approach than Convolutional Neural Network (CNN) models to handle the task and demonstrated a promising performance.

Generally, the attention module tends to exert a learnable weight on features to distinguish importance from various perspectives. According to the computing techniques and tasks, the attention modules can be categorized into two types. Hard attention module is non-differentiable and requires computation tricks \cite{rvattention,Ba2015MultipleOR}, while soft attention module is differentiable thus having wider applications for visual tasks \cite{residualatt,SEnet,cbam}. As a particular form of attention,  self-attention is applied as the core mechanism of the neural network named transformer, which is widely employed in the field of computer vision. The essential advantage of self-attention is that it focuses on the relative importance of own content rather than weighting across multiple contents as a general attention mechanism. As one of the critical roles in the self-attention mechanism, dot product multiplication among the feature of query, key, and value plays a significant role in learning self-alignment\cite{synthesizer}. Precisely, employing the dot product between a single token and all other tokens in the sequence could formulate the relative importance score.
Through pairwise dot product, the self-attention mechanism establishes a content-based retrieval process.

\begin{figure}[H]
\includegraphics[width=13.5 cm]{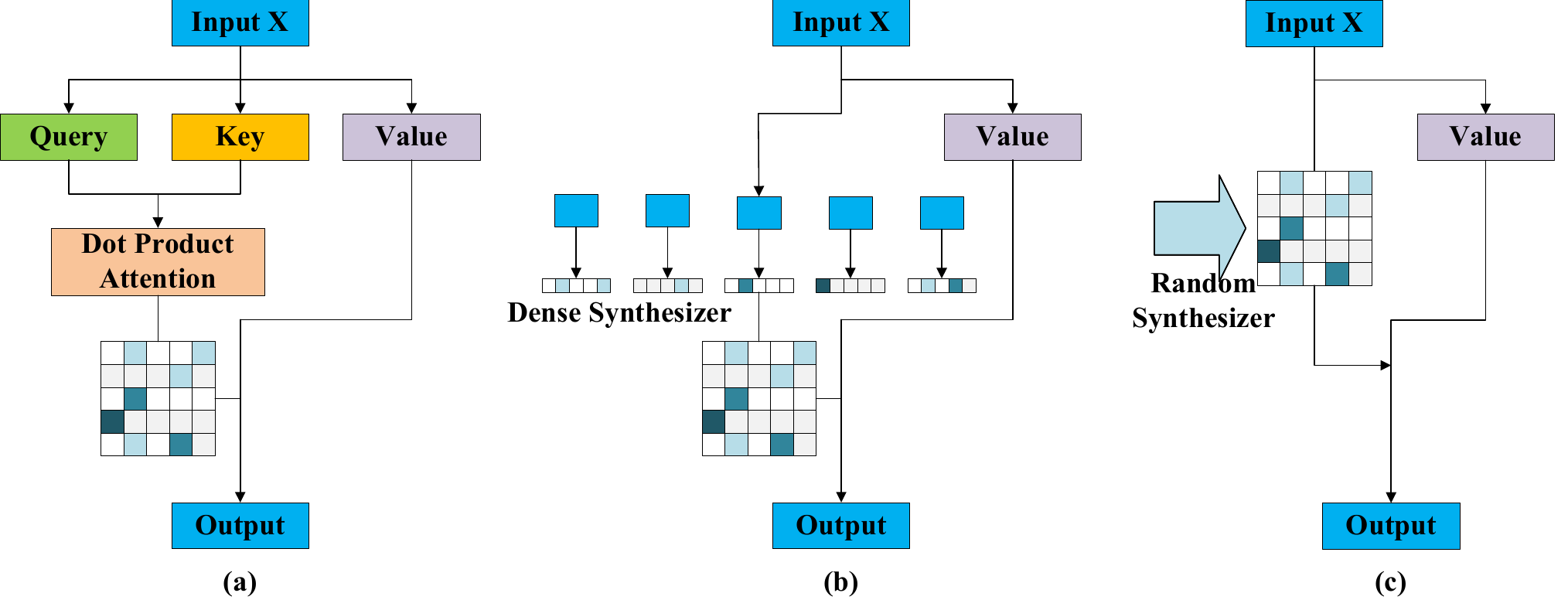}
\caption{Dot product attention VS. synthesizers for NLP\cite{synthesizer}. (\textbf{a}) Dot product based self-attention for NLP. (\textbf{b}) Dense Synthesizer for NLP. (\textbf{c}) Random Synthesizer for NLP.}\label{NLP Synthesizer}
\end{figure}

However, do we really need to use dot product in the self-attention mechanism? Is it the best choice to learn the self-alignment between the features of query and key? We can not deny that attention-based architecture is one of the most efficient models in machine learning, but it also has space to improve\cite{not2021attention}. Investigating the alternative plug-in module for current architecture could help us to have a deeper understanding of the attention mechanism. Undoubtedly, dot product multiplication has several benefits. But it also induces exhaustive and redundant computation during exerting on the features. Especially on large-scale processing data, the extra parameters and memory caused by dot product operations enormously increase the burden of the training process. Moreover, in the computer vision task, self-attention for images tends to compute the similarity scores among visual features\cite{detectdeformable,cbam,attcorrection} rather than mapping for retrieval. In order to overcome the shortages of dot production, it is necessary to search for another approach to improve the performance of the self-attention mechanism. As the result, recent findings regarding the NLP attention module have led to an alternative approach, named Synthesizer Attention\cite{synthesizer}, as shown in Figure~\ref{NLP Synthesizer}. 

\unskip

In this work, we provide a viable alternative to overcome the shortage brought from the dot-production in the standard self-attention module, in order to pursue a better performance of visual self-attention. Inspired by the NLP-like synthesizer\cite{synthesizer,app14177794}, we propose a plug-in synthesizer to replace traditional visual self-attention modules, namely, the Visual Dense Synthesizing (VDS) module.
Compared with the NLP-like synthesizer, which contains one-dimensional natural language vectors, our module applies the multi-dimensional tensor to adapt the presentation of visual features.
The module is basically composed of a tensor transformation for self-alignment between query matrix and key matrix, without computing the dot-product multiplication among the feature of query-key-value.
Removing dot product operations, the VDS gains some straightforward advantages:
\begin{enumerate}
	\item Replacing dot production based computation with a learnable linear synthesizer reduces dependency on the input, thus enhancing the robustness of the model against external perturbations;
	\item The proposed linear synthesizer can be easily adapted to input data structure, without reshaping the raw input. Thus, it is easier to preserve the underlying structure of feature maps;
	\item Since the alignment transformations in the synthesizer are computed upon various dimensions of feature maps, the attention is produced simultaneously in both channel and spacial domain.
\end{enumerate}

To further simplify the linear transformations in synthesizing, derived from the regular VDS module, the paper further proposes a series of plug-and-play modules, namely, Synthesizing Tensor Transformations (STT). The STT series are finally applied for the image classification task and image caption task. According to the result of our typical classification experiment, the STT series demonstrates that it is not only a viable alternative of the traditional self-attention module but also has better robustness. 

\section{Related Work}
\subsection{Self-Attention Mechanism}

The self-attention attracts considerable interest due to its versatile application. For CNN-based models, self-attention mechanism has been used in many modules,
such as extra re-weighting modules for channels \cite{gatherexcite,SEnet,residualatt}, jointly spatial and channel attention module \cite{chen2017sca_cvpr,cbam,dualatt},
or remold convolution operation with self-attention \cite{bello2019attention_augment_iccv,standalone,Cordonnier2020On_relation_iclr,zhao2020exploring_CVPR}.

Another line of works arrange self-attention to be a component in a pipeline for downstream tasks, 
such as augmentation of feature maps for classification \cite{bello2019attention_augment_iccv, Maurcio2023ComparingVT}, object detection \cite{detectdeformable,relationOD,endtoendOD}, segmentation \cite{axialsegmentation}, image captioning \cite{app131911103} and depth estimation\cite{tong2024robust}. 
However, it is difficult to directly apply the self-attention mechanism to pixel-wise data. 
\cite{imtransformer} restricted self-attention convolution within local neighborhoods for query pixels,
and \cite{axialsegmentation,axialtransformer} restricted computation along individual axes.
\cite{igpt} reduced image resolution and color space before transformer, and \cite{ViT} directly applied transformer on image sequence patches.

To reduce computational costs, some of the work adjust the structure of attention. 
\cite{ISSA} divided and permuted the feature map thus smaller attention map would be conquered and permuted back. 
\cite{CCnet} focused on the criss-cross around the key points on the feature of query, key and value.
\cite{a2nets} adopted double attention to re-allocate features for channels and the total workload shrinks after feature gathering.
Differently, another way is to reduce the length at certain dimension \cite{linformer}, 
or divide computation by locality sensitive hashing \cite{reformer}.

\subsection{Dot-product in Self-attention}

The regular dot-product operator based self-attention serves as a basic building block of vision tasks \cite{bello2019attention_augment_iccv,standalone,Cordonnier2020On_relation_iclr,zhao2020exploring_CVPR}.

\begin{figure}[H]
\centering
\includegraphics[width=8.5 cm]{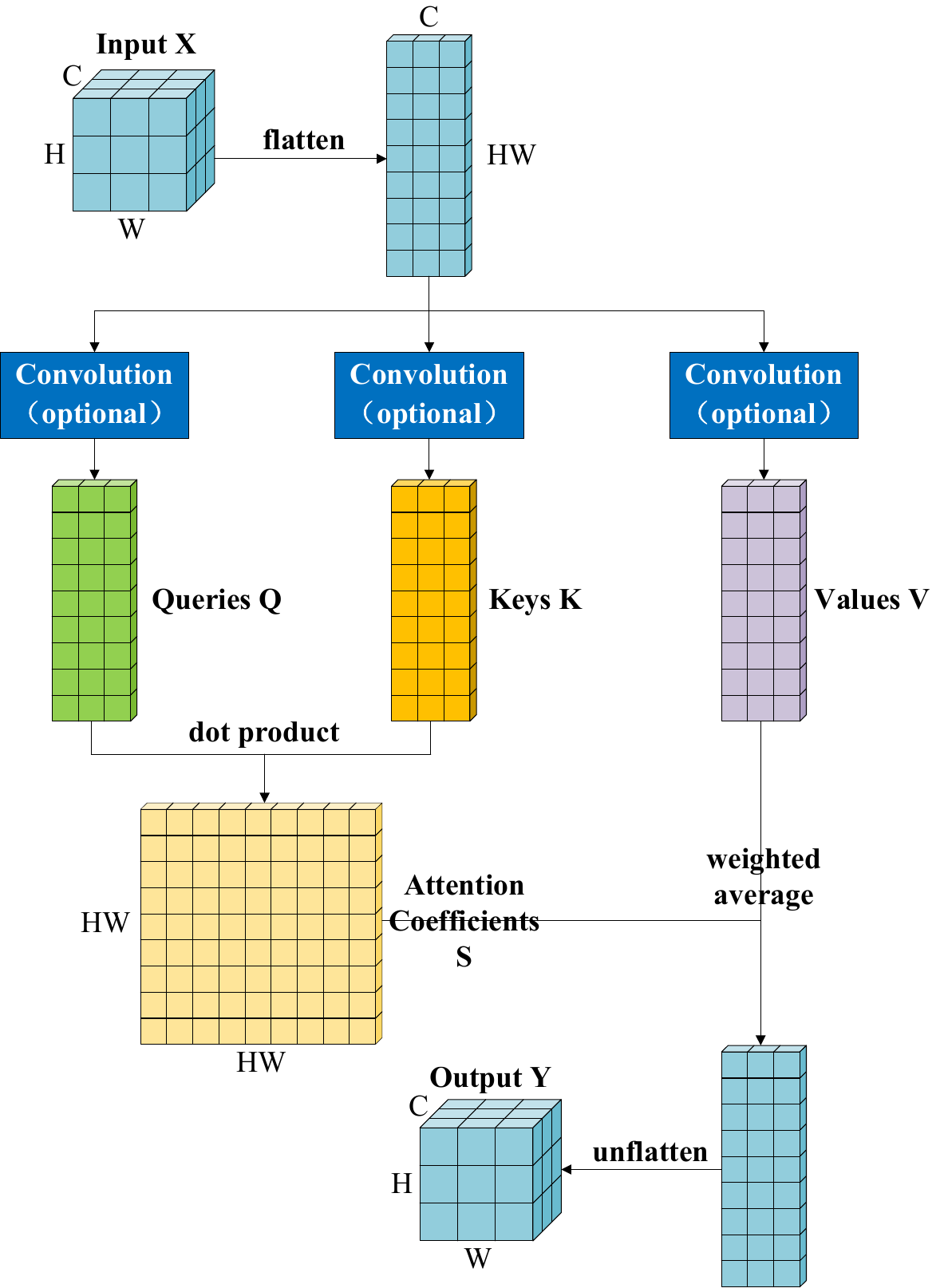}
\caption{\centering{Dot-product visual self-attention.}\label{visual_self-attention}}
\end{figure}

As is illustrated in Figure~\ref{visual_self-attention}, 
let $\mathcal{X} \in \mathbb{R}^{H \times W \times C}$ be an input tensor with the height $H$, the width $W$, the number of channels $C$.
Before being fed into the self-attention block, the 3D tensor $\mathcal{X}$ is first flattened to a 2D matrix $X \in \mathbb{R}^{HW \times C}$.
Then input $X$ is projected to corresponding representations Queries $Q \in \mathbb{R}^{HW \times C}$, Keys $K \in \mathbb{R}^{HW \times C}$ and Values $V \in \mathbb{R}^{HW \times C}$ with:
\begin{equation}
\begin{aligned}
    Q &=  \phi_{Q}(X),\\
    K &=  \phi_{K}(X),\\
    V &=  \phi_{V}(X),
\end{aligned}
\end{equation}
where $\phi_{Q}, \phi_{K}, \phi_{V}$ are trainable transformations such as linear mappings or convolutions, and have matching output dimensions. 

Take linear mapping as an example, $\phi_{Q}, \phi_{K}, \phi_{V}$ can be represented with 2D matrices 
$W_{Q} \in \mathbb{R}^{C \times d}$, $W_{K} \in \mathbb{R}^{C \times d}$ and $W_{V} \in \mathbb{R}^{ C\times d}$, respectively. 
$d$ is the desired output dimension of $Q$, $K$, $V$. 
The corresponding representations of Queries $Q$, Keys $K$ and Values $V$ can be computed by:
$$
    \begin{aligned}
        Q &=  X W_{Q},\\ 
        K &=  X W_{K},\\ 
        V &=  X W_{V}
    \end{aligned}
$$

The attention coefficients $S \in \mathbb{R}^{HW \times HW}$ can be obtained by dot product operation:
\begin{equation}
    \label{eq_dot_product_attn_coefficients}
    S = \text{Softmax}\left( \dfrac{Q K^{\top}}{\sqrt{d}}  \right),
\end{equation}
where $S_{ij}$ measures the similarity between the $i^{\text{th}}$ row of $Q$ and the $j^{\text{th}}$ row of $K$. 
Finally, the output $Y \in \mathbb{R}^{HW \times C}$ is the weighted average over Values $V$ with coefficients $S$: 
\begin{equation}
    Y = \text{Attention} (Q, K, V) =  S V
\end{equation}

\unskip

\section{Materials and Methods}

In order to introduce our Synthesizing Tensor Transformation, we first give a fundamental model of tensor synthesizer. And then, the specific tensor synthesizers would be provided as (a) Tensor Dense Synthesizer, (b) Tensor Random Synthesizer, and (c) Tensor Factorized Synthesizer.

\subsection{Preliminaries}
\paragraph{\textbf{Basic Tensor Synthesizer}}
In the case of regular visual transformation architecture\cite{ViT}, the tensor input $\mathcal{X} \in \mathbb{R}^{ H \times W \times C}$ is reshaped into 2D matrix $X \in \mathbb{R}^{HW \times C}$ and the dot-product self-attention could be interpreted as producing a dimension alignment from $\mathbb{R}^{HW\times C}$ to $\mathbb{R}^{HW\times HW}$, as is shown in Figure~\ref{visual_self-attention}.

However, the flattening of input tensor may result in the loss of spatial information and a prohibitively expensive computation when the dimension is high.
As a result, in order to directly process the tensor input $\mathcal{X} \in \mathbb{R}^{H \times W \times C}$, we propose to synthesize tensor transformation for replacing the dot-product self-attention in this section.

Formally, let $\mathcal{Z}\in \mathbb{R}^{ H \times W \times d}$ be the projected features of 
$\mathcal{X} \in \mathbb{R}^{ H \times W \times C}$ via the mapping function $ \mathcal{F} (\cdot) $, where $C$ and $d$ are the number of channels.
\begin{equation}
    \mathcal{Z} = \mathcal{F}(\mathcal{X})
\end{equation}

\begin{figure}[H]
\centering
\includegraphics[width=7.5 cm]{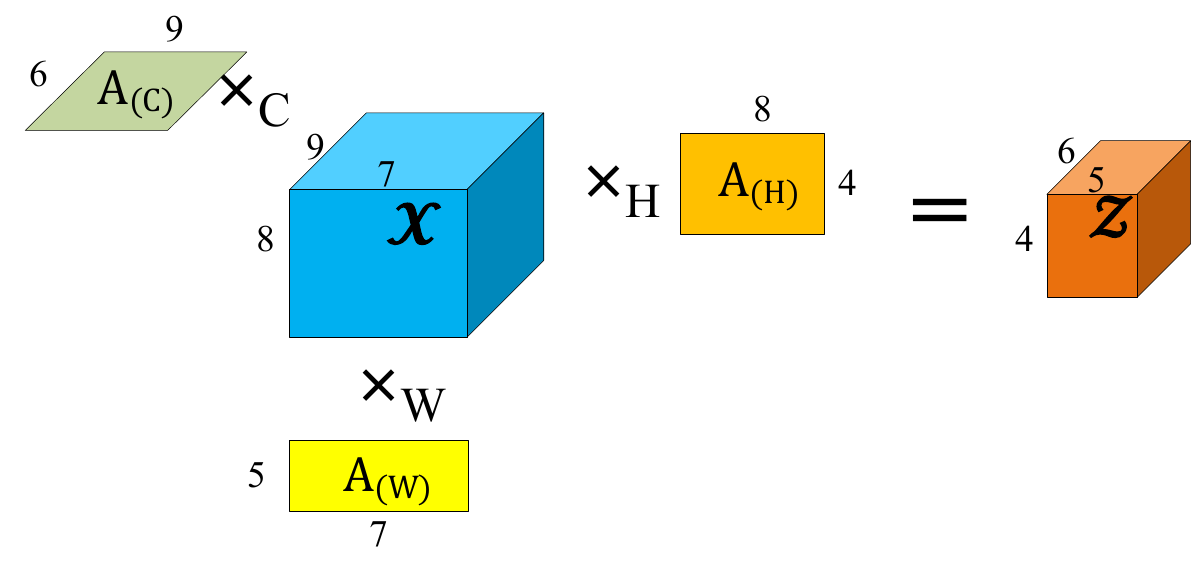}
\caption{
This figure visualizes an example of the $n-$mode product. The third-order tensor $\mathcal{X} \in \mathbb{R}^{8\times7\times9}$ is mapped into $\mathcal{Z} \in \mathbb{R}^{4\times5\times6}$ with a 3-mode product using matrices	$\mathbf{A}_{(H)} \in \mathbb{R}^{4\times8}$, $\mathbf{A}_{(W)} \in \mathbb{R}^{5\times7}$,	$\mathbf{A}_{(C)} \in \mathbb{R}^{6\times9}$. \label{fig_tensor_example}}
\end{figure}

Intuitively, $\mathcal{F} (\cdot)$ could be a 3-mode tensor product. In detail, the mapping function $\mathcal{F}(\cdot)$ could be implemented as a three-layered feed-forward network, as follow: 
\begin{equation}
	\label{eq_mapping_function}
	\mathcal{F}(\mathcal{X}) =  \mathcal{X} \times_{H} \mathbf{A}_{(H)}  \times_{W} \mathbf{A}_{(W)} \times_{C}  \mathbf{A}_{(C)},
\end{equation}
with $\mathbf{A}_{(H)}$, $\mathbf{A}_{(W)}$ , and $\mathbf{A}_{(C)}$ are three matrices. For the convenience of understanding the tensor product, Figure~\ref{fig_tensor_example} gives an example of n-mode production. For a detailed introduction of tensor products, we refer the interested readers to  \cite{de2000multilinear_siam, van2000ubiquitous_jcam, tensorproduct}.

\unskip

\begin{Definition}[$n-$mode product]
Given a tensor $\mathcal{X}\in\mathbb{R}^{I_{1}\times I_{2}\times \cdots \times I_{N}}$ and a matrix $\textbf{A}_{n}\in\mathbb{R}^{J_{n} \times I_{n}}$, the n-mode product is denoted by:
\begin{equation}
    \mathcal{X} \times_{n} \textbf{A}_{n},
\end{equation}
and results in an $I_{1}\times I_{2}\times \cdots \times I_{n-1} \times J_{n}\times  I_{n+1} \times \cdots \times I_{N}$ tensor. The entries of this tensor are defined as:
\begin{equation}
    \left(\mathcal{X} \times_{n} \mathbf{A}_{n}\right)_{i_{1} i_{2} \cdots i_{n-1} j_{n} i_{n+1} \cdots i_{N}}=\sum_{i_{n}=1}^{I_{n}} x_{i_{1} i_{2} \cdots i_{n-1} i_{n} i_{n+1} \cdots i_{N}} \cdot a_{j_{n} i_{n}},
\end{equation}
for all $j_{n}=1,\cdots,J_{n}$.
\end{Definition}

Besides, in order to obtain an appropriate size of the matrix, a method named \textit{n-mode matrix unfolding} is required. 
Depending on this method, the multiplication is conducted between a k-order tensor $\mathcal{X}$ along all its modes and matrices $A_{1},\cdots,A_{k}$\cite{tensorproduct} and the layered feed-forward network could be built as follow: 
\begin{equation} 
\mathcal{Y}=\mathcal{X} \times_{1} A_{1} \times_{2} A_{2} \times_{3} \cdots \times_{k} A_{k} 
\end{equation} 

\begin{figure}[H]
\centering
\includegraphics[width=10.5 cm]{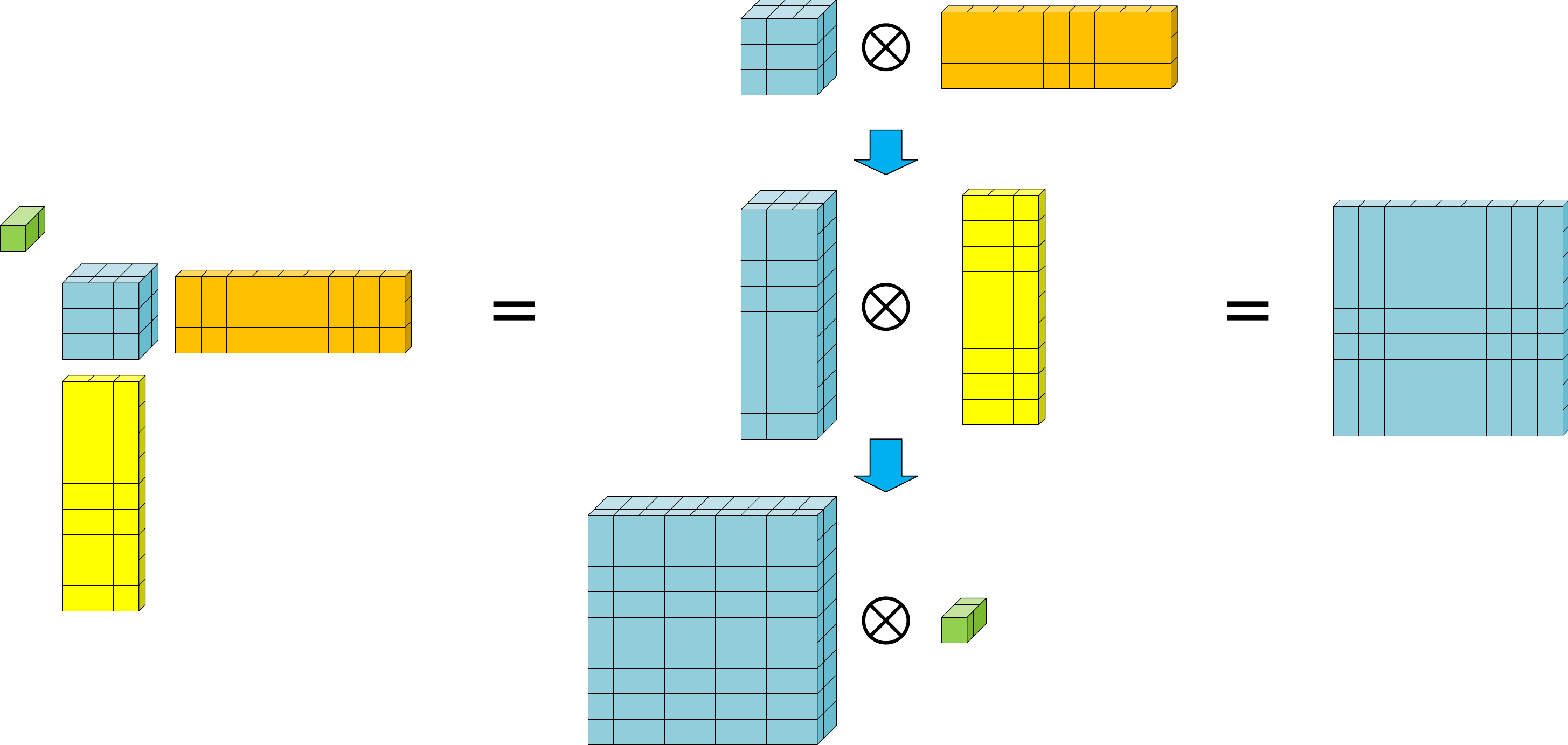}
\caption{Illustration for tensor multiplication with Kronecker procduct operator. In this case, a three-dimensional tensor is first expanded along its vertical dimension. Then, the horizontal dimension is expanded. After that, the third dimension is reduced to get a two-dimensional matrix. \label{fig_tensor_production}}
\end{figure}

It can be rewritten as the following linear system: 
\begin{equation}
\operatorname{vec}(\mathcal{Y})=\left(A_{k} \otimes \cdots \otimes A_{2} \otimes A_{1}\right) \operatorname{vec}(\mathcal{X}), 
\end{equation}
where $vec(\cdot)$ means the vector space isomorphism: $\mathbb{R}^{I_{1}\times I_{2}\times \cdots \times I_{k}} \rightarrow \mathbb{R}^{I_{1} I_{2} \cdots I_{k}}$, which denotes the operator that unfolds a tensor along the last dimension\cite{tensorproduct}. 
Therein, $\otimes$ denotes the Kronecker product operator. with which the Kronecker product of A and B is defined by:
\begin{equation}
    \textbf{A} \otimes \textbf{B} = \begin{bmatrix}
        A_{11}\textbf{B}  &\cdots  &A_{1k_{1}}\textbf{B}\\
        \vdots  &\ddots  &\vdots\\
        A_{a1}\textbf{B}  &\cdots  &A_{ak_{1}}\textbf{B}\\
    \end{bmatrix}, 
\end{equation}
with $A_{ij}$ being an entry in \textbf{A}. More details about the Kronecker product could be found in \cite{van2000ubiquitous_jcam}. Therefore, the mapping $\mathcal{F}(\cdot)$ could be presented as the tensor multiplication as the Eq.~\eqref{eq_mapping_function}. Figure \ref{fig_tensor_production} shows the process of the tensor multiplication with Kronecker product operator.

\unskip

\subsection{Synthesizer Tensor Model}
\subsubsection{Tensor Dense Synthesizer}

To achieve mentioned dimension alignment on raw tensor $\mathcal{X}$ without the dot-product self-attention mechanism, 
we define a tensor transformation to compute attention coefficients by replacing the dot-product self-attention.

\begin{Definition}[Tensor transformation] 
Given the input tensor $\mathcal{X} \in \mathbb{R}^{H \times W \times C}$, the attention coefficients defined in Eq.~\eqref{eq_dot_product_attn_coefficients} could be computed with n-mode production: 
\begin{align}
	\label{eq_attention_STT_trans}
	Z & = \mathcal{X} \times_{1} \mathbf{A}_{(H)}^{(l)} \times_{2} \mathbf{A}_{(W)}^{(l)} \times_{3} \mathbf{A}_{(C)}^{(l)}, \\
	\label{eq_attention_STT_cofficients}
	S & = \mathrm{Softmax} \big(Z \big),
\end{align}
with $\mathbf{A}_{(H)}^{(l)} \in \mathbb{R}^{HW\times H}$, $\mathbf{A}_{(W)}^{(l)} \in \mathbb{R}^{HW\times W}$, 
$\mathbf{A}_{(C)}^{(l)} \in \mathbb{R}^{1\times C}$, $Z, S \in  \mathbb{R}^{HW\times HW}$, and $l$ denotes the layer number.
\end{Definition}

\begin{figure}[H]
\begin{adjustwidth}{-\extralength}{0cm}
\centering
\includegraphics[width=16.5 cm]{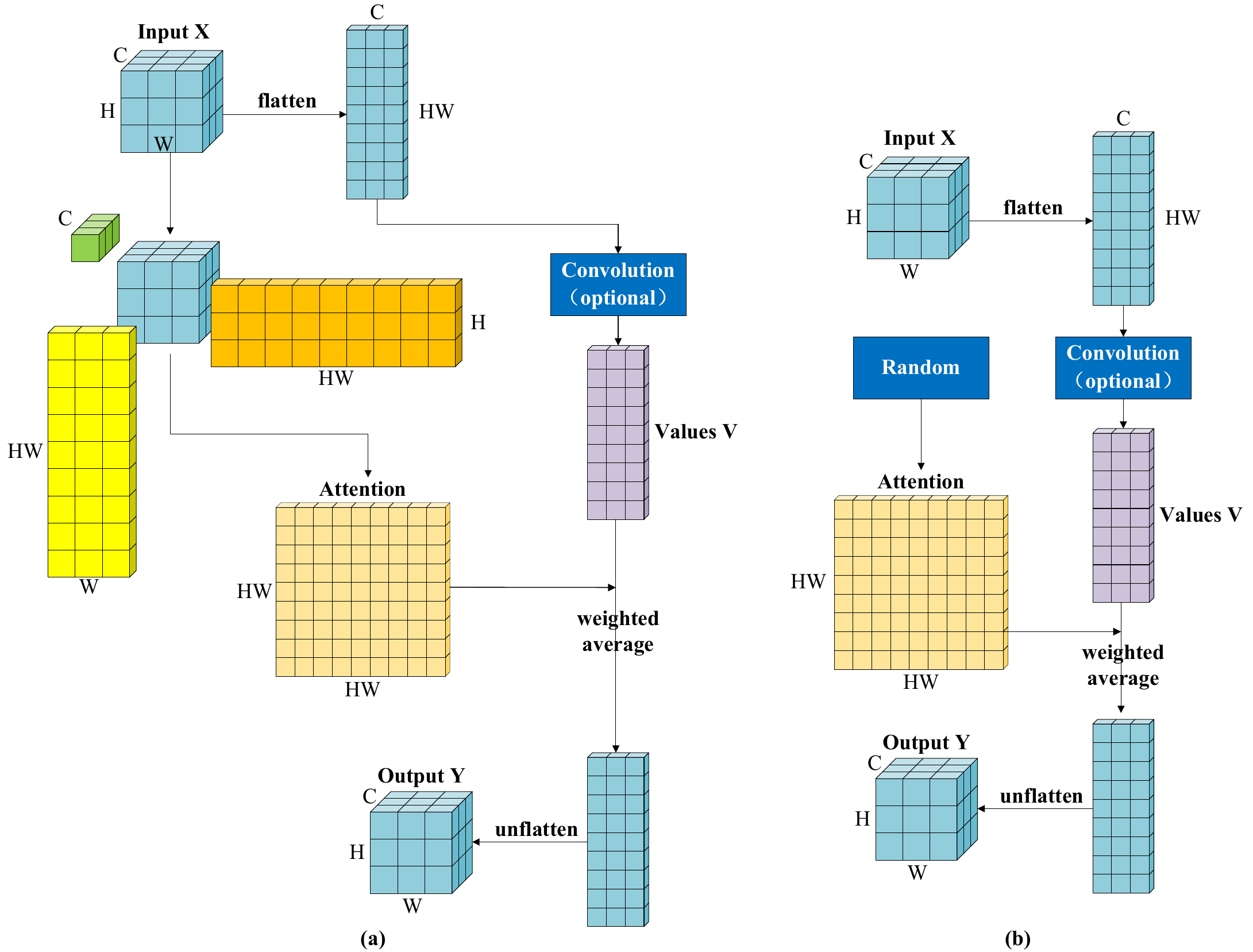}
\caption{Dense synthesizers and random synthesizer for vision tasks. (\textbf{a}) Tensor Dense Synthesizer for vision tasks. (\textbf{b}) Tensor Random Synthesizer for vision tasks. \label{fig_dense_random}}
\end{adjustwidth}
\end{figure}

The combination of Eq.~\eqref{eq_attention_STT_trans} and Eq.~\eqref{eq_attention_STT_cofficients} is referred to the VDS module, which eliminates the dot product altogether by replacing $QK^\top$ in standard self-attention with the synthesizing function defined in Eq.~\eqref{eq_attention_STT_trans}, as shown in Figure~\ref{fig_dense_random}(\textbf{a}). 
Then, the output is computed by: 
\begin{equation}
	\label{eq_attention_STT_output}
	Y = \text{Synthesizer} (Q, V) = SV
\end{equation}

\unskip

\begin{figure}[H]
\centering
\includegraphics[width=9.5 cm]{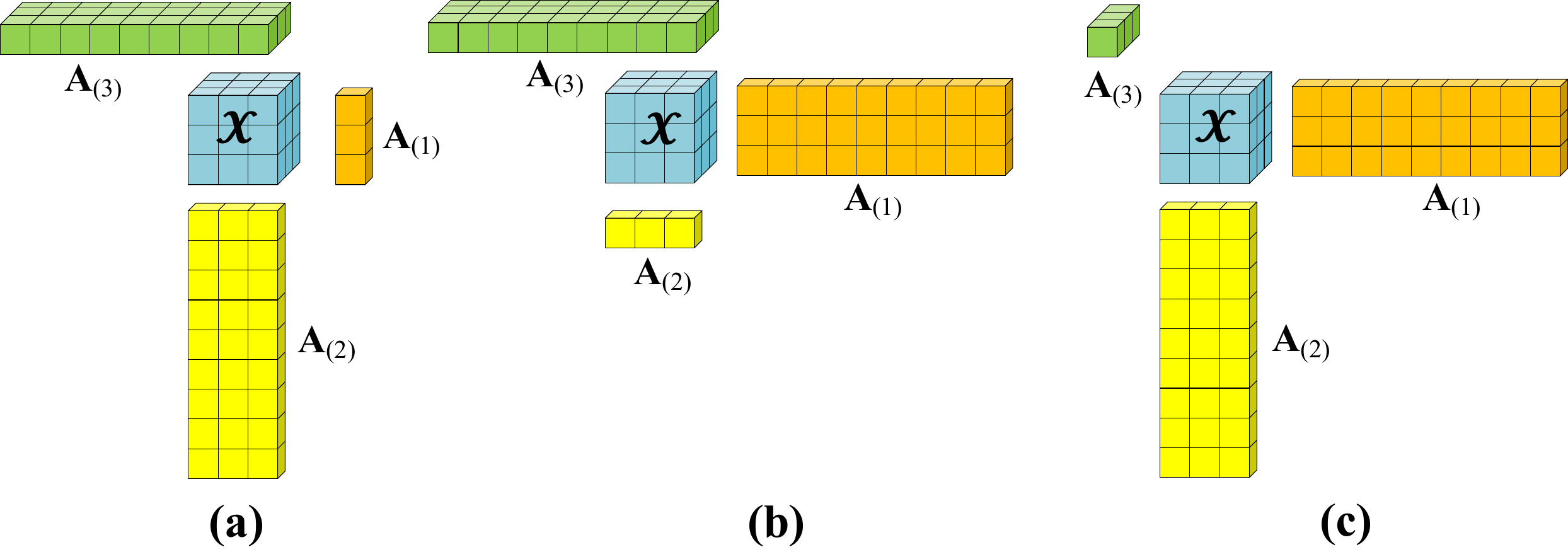}
\caption{The same dimension alignment can be achieved with three different transformations: (\textbf{a}) reduce the first dimension, (\textbf{b}) reduce the second dimension or (\textbf{c}) reduce the third dimension, while expand the other two dimensions. \label{fig_decomposition}}
\end{figure}

However, to achieve dimension alignment on input tensor $\mathcal{X}$ with Eq.~\eqref{eq_attention_STT_trans}, the output tensor $Z \in \mathbb{R}^{HW \times HW}$ could be assigned the dimension with three different transformations, as shown in Figure~\ref{fig_decomposition}. 
Therein, $Z_H, Z_W, Z_C $ are computed as follows: 
$$Z_H = \mathcal{X} \times_{1} \mathbf{A}_{(1)} \times_{2} \mathbf{A}_{(2)} \times_{3} \mathbf{A}_{(3)}, $$ 
with $\mathbf{A}_{(1)} \in  \mathbb{R}^{1\times H}$, $\mathbf{A}_{(2)} \in \mathbb{R}^{HW\times W}$, 
$\mathbf{A}_{(3)} \in \mathbb{R}^{HW\times C}$. 
$$Z_W = \mathcal{X} \times_{1} \mathbf{A}_{(1)}  \times_{2} \mathbf{A}_{(2)} \times_{3}  \mathbf{A}_{(3)}, $$
with $\mathbf{A}_{(1)} \in  \mathbb{R}^{HW\times H}$, $\mathbf{A}_{(2)} \in \mathbb{R}^{1\times W}$, 
$\mathbf{A}_{(3)} \in \mathbb{R}^{HW \times C}$,
$$Z_C = \mathcal{X} \times_{1} \mathbf{A}_{(1)} \times_{2} \mathbf{A}_{(2)} \times_{3} \mathbf{A}_{(3)}, $$
with $\mathbf{A}_{(1)} \in \mathbb{R}^{HW \times H}$, $\mathbf{A}_{(2)} \in \mathbb{R}^{HW\times W}$, 
$\mathbf{A}_{(3)} \in \mathbb{R}^{1\times C}$. 

\unskip

It is difficult to determine the dimensions of $\mathbf{A}_{(H)}^{(l)}$, $\mathbf{A}_{(W)}^{(l)} $, $\mathbf{A}_{(C)}^{(l)} $, i.e.,how to assign the dimensions $HW, H, W, 1$ to three transformations. We note that all of the proposed synthetic attention variants can be mixed in an additive fashion. As a result, Eq.~\eqref{eq_attention_STT_output} can be expressed as:
\begin{equation}
\label{eq_attention_STT_output_3direction} 
Y  = \text{Softmax} \big(Z_H \big)V  + \text{Softmax} \big(Z_W \big)V + \text{Softmax} \big(Z_C \big)V 
\end{equation}

\subsubsection{Tensor Random Synthesizer}

In the model of Tensor Dense Synthesizer, the features of input $\mathcal{X}$ are projected from $\mathbb{R}^{ H \times W \times C}$ to $\mathbb{R}^{HW \times HW}$ via the 3-mode tensor product. Intuitively, the synthesizer implements the conditioning on each dimension independently. In contrast, the Tensor Random Synthesizer utilizes the random value as initialized attention weights, as shown in Figure~\ref{fig_dense_random}(\textbf{b}). Then, the attention weights could either follow the training process to update, or keep fixed. 

Since the synthesizer can be regarded as a self-alignment of dimensions, we can make an attempt to set $Z$ as a random matrix $R$ and the Random Synthesizer is defined as: 
\begin{equation}
    Y = \text{Synthesizer} (Q, V) = \text{Softmax} \big( R \big) V,
\end{equation}
with $R \in \mathbb{R}^{HW\times HW}$ being a randomly initialized matrix. The attempt of utilizing the random initial matrix identifies the idea that it is unnecessary to rely on any information from the individual input feature. Therefore, the Tensor Random Synthesizer is tending to learn the task-specific alignment through the training process. 

Compared with the token-token self-attention, the synthesizers are able to handle the input tensors with longer sequences or higher dimensionality. In our work, the factorized approach will further reduce the computation, and we will introduce it in the next section. 

\subsection{Multiple-level Factorized Tensor Synthesizer}

We replace the dot-production in the standard self-attention module with our synthesizer module in the Tensor Dense Synthesizer. However, the synthesizer is too burdensome to learn when the size of parameters is too large. As shown in Eq.~\eqref{eq_attention_STT_trans}, the basic 3-mode tensor synthesizer contains several parameters 
$\mathbf{A}_{(H)}^{(l)} \in \mathbb{R}^{HW\times H}$, $\mathbf{A}_{(W)}^{(l)} \in \mathbb{R}^{HW\times W}$, 
$\mathbf{A}_{(C)}^{(l)} \in \mathbb{R}^{1\times C}$. 
Therefore, there are several factorized tensor synthesizer models introduced. 

\subsubsection{Tensor Dense Factorized Synthesizer}

\begin{figure}[H]
\centering
\includegraphics[width=10.5 cm]{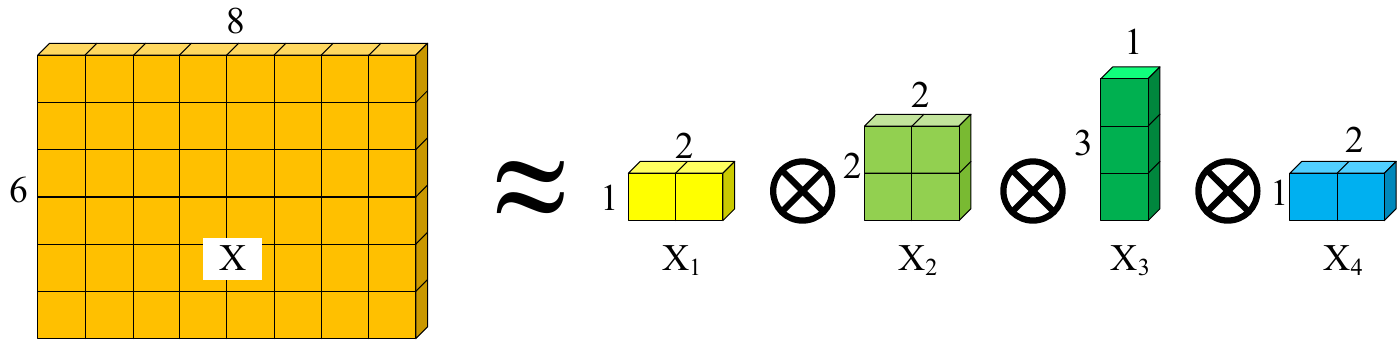}
\caption{An example of Kronecker Decomposition. As the inverse process of Kronecker production, $X \in \mathbb{R}^{6\times8}$ could be decomposed into $X_{1} \in \mathbb{R}^{1\times2}$, $X_{2} \in \mathbb{R}^{2\times2}$, $X_{3} \in \mathbb{R}^{3\times1}$ and $X_{4} \in \mathbb{R}^{1\times2}$, where $6=1\times2\times3\times1$ and $8=2\times2\times1\times2$. \label{fig_Kronecker_Decomposition}}
\end{figure}

In order to reduce the size of parameters, partially avoiding the over fitting while $HW$ being huge, the parameter matrices within the Kronecker product could be decomposed\cite{van2000ubiquitous_jcam}, 
as shown in Figure~\ref{fig_Kronecker_Decomposition}. 
The Kronecker decomposition is the inverse process of Kronecker product operation.

\unskip

For example, $\mathbf{A}_{(H)}^{(l)} $ could be approximated by $N$ small matrices as: 
\begin{equation}
\label{eq_A_(H)}
\mathbf{A}_{(H)}^{(l)} \approx \mathcal{D}\Big(\mathbf{A}_{(H)}^{(l)}\Big) = \mathbf{A}_{(H)_1}^{(l)} \otimes \mathbf{A}_{(H)_2}^{(l)} \otimes \cdots \otimes \mathbf{A}_{(H)_N}^{(l)},
\end{equation}
where $\mathcal{D}(\cdot)$ represents Kronecker decomposition, $\mathbf{A}_{(H)_i}^{(l)} \in \mathbb{R}^{\alpha_{i} \times \beta_{i}}$ for $i=1,2,\cdots,N$, $HW = \alpha_{1}\alpha_{2}\cdots \alpha_{N}, H = \beta_{1}\beta_{2}\cdots \beta_{N}$. By using the Kronecker decomposition, $\mathbf{A}_{(H)}^{(l)} $, $\mathbf{A}_{(W)}^{(l)} $, 
$\mathbf{A}_{(C)}^{(l)} $ can be easily factorized to several very small matrices. The Tensor Dense Synthesizer in Eq.\eqref{eq_attention_STT_output} could be reformulated as Tensor Dense Factorized Synthesizer, which is defined as:
\begin{equation}
\label{eq_dense_factor}
\begin{aligned}
\mathcal{Z} &= \mathcal{X} \times_{H} \mathcal{D}_{H}\Big(\mathbf{A}_{(H)}^{(l)}\Big) \times_{W} \mathcal{D}_{W}\Big(\mathbf{A}_{(W)}^{(l)}\Big) \times_{C} \mathcal{D}_{C}\Big(\mathbf{A}_{(C)}^{(l)}\Big),\\
\mathcal{S} &= \text{Softmax}(\mathcal{Z}),\\
Y &= \text{Synthesizer}(Q,V) = \mathcal{S}V,     
\end{aligned}
\end{equation}
where $\mathcal{D}(\cdot)$ decomposes the weight matrices. 

\subsubsection{Tensor Random Factorized Synthesizer}

Similarly, the factorized synthesizer could also be applied into Tensor Random Synthesizer, as shown in Figure~\ref{fig_Kronecker_Decomposition}. The idea is to decompose the random coefficient matrix $R \in \mathbb{R}^{HW \times HW}$ into several low rank matrices $R_{i} \in \mathbb{R}^{\alpha_{i} \times \beta_{i}}, i=1,2,\cdots, N$: 
\begin{equation}
\label{eq_random_factor}
\begin{aligned}
\mathcal{R} &= R_{1} \otimes R_{2} \otimes \cdots \otimes R_{N},\\
Y &= \text{Synthesizer}(Q,V) = \text{Softmax}(\mathcal{R}) V,
\end{aligned}
\end{equation}
where $\alpha_{1}\alpha_{2}\cdots \alpha_{N}=HW$, $\beta_{1}\beta_{2}\cdots \beta_{N}=HW$, and $\sum_{i=1}^N{\alpha_{i}\beta_{i}} << HW\times HW$. With the reduced size of parameters, it is able to avoid the over-fitting problem. 

\subsubsection{Mixture Tensor Synthesizer}

With the above four kinds of tensor synthesizer models been introduced, a compounded model could be proposed by mixing the synthesizers in additive fashion. For instance:
\begin{equation}
\label{eq_mix} 
Y = \text{Softmax} (\theta_{1}S_{1}(\mathcal{X})+\cdots+\theta_{M}S_{M}(\mathcal{X})) V,
\end{equation}
where $\theta_{i}$ are the learnable weights, and $\sum_{i=1}^{M} \theta_{i}=1$. $S_{i}$ are the different synthesizer models. To be specific, the Mixture Tensor Synthesizer in the experiments is composed of Tensor Random Factorized Synthesizer and Tensor Dense Synthesizer, which could be represented as follow:
\begin{equation}
\label{eq_mix2}
Y = \text{Softmax} (\theta_{1}\mathcal{S} + \theta_{2}\mathcal{R}) V,
\end{equation}
where $\mathcal{S}$ and $\mathcal{R}$ are defined in Eq.~\eqref{eq_dense_factor} and Eq.~\eqref{eq_random_factor}, respectively. By employing more than one kind of synthesizer, it is possible to achieve better performance. 

\section{Results}

In this section, the performance of our proposed {STT} module series on vision tasks is thoroughly investigated on image classification and image captioning. 
The experiments focus on the performance of the robustness of the model. 
For convenience, we abbreviate the proposed {STT} series and comparison methods. The detail of information are listed in Table~\ref{tab:abbrl}. Matrix Dense Synthesizer being the basic synthesizer proposed in \cite{synthesizer}, and Tensor Dense Synthesizer being our basic tensor synthesizer for vision tasks, Tensor Height Synthesizer being the tensor synthesizer with an attention map full-rank in height, and Mixture Tensor Synthesizer being the synthesizers mixed in additive fashion.

\begin{table}[H]
\caption{Abbreviation of Model Zoo.\label{tab:abbrl}}
\newcolumntype{C}{>{\centering\arraybackslash}X}
\begin{tabularx}{\textwidth}{C|C}
    \toprule
    \textbf{ABBR} * & \textbf{Combination}\\
    \midrule
    None   & Convolution Baseline \\
    CT     & CNN + Transformer \cite{vaswani2017attention_NeurIPS} \\
    CL     & CNN + Linformer \cite{linformer} \\
    \hline
    SD     & CNN + Matrix Dense Synthesizer \cite{synthesizer}\\
    STT    & CNN + Tensor  Dense Synthesizer\\
    SR     & CNN + Tensor Random Synthesizer\\
    STTH   & CNN + Tensor Height Synthesizer\\
    STTW   & CNN + Tensor Width Synthesizer\\
    FSD    & CNN + Tensor Dense Factorized Synthesizer\\
    FSR    & CNN + Tensor Random Factorized Synthesizer\\
    MS     & CNN + Mixture Tensor Synthesizers\\
    \bottomrule
\end{tabularx}
\noindent{\footnotesize{* ABBR indicates abbreviations of our proposed STT series.}}  
\end{table}

\subsection{Image Classification}
\subsubsection{Implementation Details}

STT series are validated on the benchmark image dataset CIFAR10\cite{cifar100} for classification. The CIFAR10 is composed by 10 classes, and it has 50000 images for training, 10000 for testing.
To validate the functionality of the STT series as attention modules applied to a deep learning framework, we employed three different perturbations to test the model robustness of above all approaches. The applied perturbations include Gaussian noise, image rotations and flipping.

\begin{enumerate}
\item Gaussian noise: random noise is added on every pixel of the input and we want to test whether the attention computed on the global scale out performs the split ones. 
\item Rotations and flips: once the rotations or flips are exerted, the order among channels would change correspondingly, while STT series treat the feature map on the whole.
\end{enumerate}
To ensure fair comparisons with traditional self-attention modules, we use a two-layer CNN as the baseline and evaluate performance based on classification accuracy. Besides, we select Transformer\cite{vaswani2017attention_NeurIPS}, Linformer\cite{linformer}, and Matrix Dense Synthesizer\cite{synthesizer} to test their robustness in the same baseline structure. 

The Transformer and Linformer modules employ the dot-product to compute the similarity between token to token. And Matrix Dense Synthesizer also applies the synthesizer mechanism to replace the dot production, but it doesn't leverage the basic tensor synthesizer in their module. Through the comparison, we could prove our STT series satisfy the requirement to become a valuable alternative to the dot-product-based self-attention model. Moreover, it has better robustness performance. 
\subsubsection{Quantitative Analysis on Robust Classification}
According to our proposed STT series, one of the advantages is that they possess strong robustness against external perturbations. The reason is that they calculate attention maps without splitting the feature maps. Thus, the baseline model reduces the reliance on the input, and the underlying structure of the feature map has more possibility to preserved. 

Our STT series has a more negligible effect from external perturbation on the raw input than the dot-production-based self-attention modules. Hence, to evaluate the robustness validation for our proposed STT series, we conduct the image classification task to compare the STT series with traditional self-attention modules. 

\paragraph{\textbf{Gaussian Noise}}

As a common statistical noise caused by sensor during sampling or transmission, Gaussian noise is a widespread type of disturbance for digital images.
Varied levels of Gaussian noises are utilized to corrupt the images for model robustness evaluation, with detailed settings and evaluation results provided in Table~\ref{tab:noisesmall}.

\begin{table}[H]
\caption{Classification Accuracy on Images with Relatively Small Gaussian Noise.\label{tab:noisesmall}}
    \begin{adjustwidth}{-\extralength}{0cm}
        \newcolumntype{C}{>{\centering\arraybackslash}X}
        \begin{tabularx}{\fulllength}{C|CCCCCCCCCC}
        \toprule
        \textbf{Noise}   &\textbf{0.01}   &\textbf{0.02}	&\textbf{0.03}   &\textbf{0.05}	&\textbf{0.04}	&\textbf{0.06}   &\textbf{0.07}   &\textbf{0.08}   &\textbf{0.09}   &\textbf{0.1}\\
        \midrule
        None    &45.96	&34.35	&27.8	&22.83	&20.01	&17.86	&16.23	&15.07	&14.45	&13.76\\
        CT      &47.02	&35.83	&29.21	&24.7	&21.49	&19.33	&17.41	&16.51	&15.42	&14.8\\
        CL      &41.51	&28.54	&21.69	&18.56	&16.16	&14.53	&13.57	&12.41	&12.17	&11.5\\
        \hline
        SD      &48.71	&\textbf{37.79}	&31.54	&27.3	&23.7	&22.23	&20.18	&18.92	&17.45	&17.03\\
        SR      &45.93	&34.02	&27.06	&23.01	&19.91	&17.36	&16.02	&14.93	&13.93	&13.4\\
        FSR     &46.51	&37.28	&31.69	&27.56	&24.14	&22.6	&20.62	&19.75	&18.25	&17.76\\
        FSD     &\textbf{68.9}	&37.5	&30.42	&25.89	&22.85	&20.7	&19.33	&17.91	&16.87	&16.07\\
        STT     &47.29	&36.81	&\textbf{31.98}	&\textbf{28.11}	&\textbf{25.49}	&\textbf{23.75}	&\textbf{22.39}	&\textbf{21.44}	&\textbf{20.09}	&\textbf{19.54}\\
        STTH    &44.66	&30.92	&23.26	&19.58	&17.08	&15.12	&14.04	&13.35	&12.61	&12.26\\
        STTW    &42.2	&30.21	&23.65	&19.92	&17.53	&16.41	&15.18	&14.65	&14.33	&13.64\\
        MS      &48.2	&36.8	&29.55	&25.13	&21.45	&19.34	&17.74	&16.61	&15.76	&15.26\\
        \bottomrule
        \end{tabularx}
    \end{adjustwidth}
\end{table}

As shown in Table~\ref{tab:noisesmall},
the accuracy of the classification decreases with the Gaussian noise increasing. 
From the comparison of the two series of methods, it can be concluded that as the variance in the Gaussian distribution increases, the STT series exhibits greater noise resistance than models utilizing dot-production.
When the variance of the Gaussian noise is 0.01, the FSD method of the STT series achieves the highest accuracy of 68.9, significantly outperforming other methods. As the noise variance increases, the STT method shows an outstanding advantage, achieving the best performance across all methods when the noise variance ranges from 0.03 to 0.1. 
Among the STT series, the worse performance of STTH and STTW might result from the inferior representation because the information is only gathered from one dimension of the feature map compared to the transformer. Compared with the modules employing dot-product, synthesizing modules appear higher accuracy when Gaussian noise influences them.

\paragraph{\textbf{Robustness against Flipping}}
Flipping are common in image augmentation process, which could be categorized into horizon flipping, vertical flipping, and flipping both in horizon and vertical. The robustness evaluation results against flipping are recorded in Table~\ref{tab:flip}.

\begin{table}[H]
\caption{Classification Accuracy on images with Flips, where he+ve means that both horizontal and vertical flips are applied.\label{tab:flip}}
\newcolumntype{C}{>{\centering\arraybackslash}X}
\begin{tabularx}{\textwidth}{C|CCC}
    \toprule
    \textbf{Module}  &\textbf{horizon}   &\textbf{vertical}   &\textbf{ho+ve} \\
    \midrule
    None    &64.74	&25.18	&25.08\\
    CT      &64.93	&25.52	&25.45\\
    CL      &65.31	&25.2	&25.05\\
    \hline
    SD      &65.7	&25.08	&25.21\\
    SR      &65.86	&26.06	&25.72\\
    FSR     &65.08	&24.63	&25.14\\
    FSD     &\textbf{66.28}	&26.41	&26.58\\
    MS      &65.6	&25.75	&25.99\\
    STT	    &66.09	&27.16	&27.32\\
    STTH    &66.09	&\textbf{29.29}	&\textbf{28.9}\\
    STTW    &65.77	&27.45	&27.46\\
    \bottomrule
\end{tabularx}
\end{table}

From Table~\ref{tab:flip}, we observe a significant decrease in accuracy once the vertical transformation is applied. Comparing the synthesizing modules with those employing dot-product, there is no significant difference between the two approaches. It demonstrates that the synthesizer modules are not more vulnerable under the flipping noise. Even though all the models have severe impair from vertical flipping noise, the STT series has slightly better accuracy. 

\paragraph{ \textbf{Images with Rotations}}
We apply rotations(Unit: Degrees) to the original images to evaluate the robustness with the model zoo. Detailed settings and results are recorded in Table \ref{tab:rotation}.

\begin{table}[H]
\caption{Classification Accuracy on Images with Rotations.\label{tab:rotation}}
\begin{adjustwidth}{-\extralength}{0cm}
    \newcolumntype{C}{>{\centering\arraybackslash}X}
    \begin{tabularx}{\fulllength}{C|CCCCCCCCCCCC}
    \toprule
    \textbf{Rotation}  &\textbf{30}  &\textbf{60}  &\textbf{90}  &\textbf{120}  &\textbf{150} 
 &\textbf{180}  &\textbf{210}  &\textbf{240}  &\textbf{270}  &\textbf{300}  &\textbf{330}\\
    \midrule
    None	&37.58	&26.51	&23.07	&21.12	&19.4	&19.39	&18.68	&18.89	&18.27	&17.74	&17.4\\
    CT	&36.54	&36.57	&23.27	&21.25	&19.75	&19.44	&19.7	&18.61	&18.19	&18.46	&17.75\\
    CL	&37.53	&26.86	&23.61	&21.5	&20.16	&18.61	&19.19	&17.95	&17.62	&17.95	&17.95\\
    \hline
    SD	&37.9	&\textbf{38.64}	&23.24	&21.79	&19.26	&19.45	&19.36	&18.51	&17.55	&17.6	&17.98\\
    SR 	&33.72	&24.7	&22.06	&19.88	&19	&18.17	&18.03	&17.22	&17.03	&17.23	&16.79\\
    FSR	&37.04	&26.86	&23.1	&20.69	&19.72	&18.69	&18.7	&18.27	&17.69	&17.17	&17.08\\
    FSD	&39.27	&28.85	&23.55	&22.85	&20.37	&19.85	&19.95	&18.71	&18.86	&18.34	&17.88\\
    MS	&37.91	&26.93	&22.96	&21.74	&19.61	&18.89	&18.76	&18.01	&18.06	&17.62	&17.65\\
    STT	&38.71	&38.36	&24.51	&21.8	&20.97	&20.36	&20.89	&19.93	&19.86	&19.12	&20\\
    STTH	&42.32	&30.44	&26.28	&\textbf{24.53}	&22.17	&22.18	&\textbf{22.37}	&\textbf{21.83}	&\textbf{20.85}	&20.07	&\textbf{20.16}\\
    STTW	&\textbf{45.53}	&31.72	&\textbf{27.21}	&24	&\textbf{22.39}	&\textbf{22.23}	&22.03	&21.27	&20.61	&\textbf{20.31}	&19.89\\
    \bottomrule
        \end{tabularx}
    \end{adjustwidth}
    
\end{table}

In Table \ref{tab:rotation}, 
it is obvious that the classification accuracy of the whole model zoo is impaired when rotation increases. However, the accuracy of the STT series represents a similar decreasing tendency. It proves that the STT series has the same performance as dot-production models when they are suffering the rotation noise.

Despite the STTH and STTW modules, other approaches don't reveal the resistance of rotation noise. While transformer weakens the baseline CNN structure, STTH and STTW have a relatively better accuracy performance. Based on such result, we make an assumption that the full-rank attention map to give them a better robustness against the effect of rotation. This assumption could also explain that the STT has a relatively better result, because it also has a full-rank attention map.

\subsection{Image Captioning}
\subsubsection{Implementation Details}

We report our results on  COCO-2014\cite{MSCOCO}  dataset with $82,783$ images for training, $40,504$ images for validation and $40,775$ images for testing.
The results are reported under standard evaluation protocol, where captioning metrics including full set of BLEU \cite{bleu}, along with METEOR \cite{denkowski2014meteor}, CIDEr \cite{vedantam2015cider}, ROUGE \cite{lin2004rouge}. 
Most of them  are based on $n$-gram to evaluate the quality of the generated sentences with ground truths, and they complement with each other to yield fairer results.

\subsubsection{Quantitative Analysis of Image Captioning}

The performance quantified by the seven metrics is summarized in Table~\ref{tab:imagecapresult}, from which we can see that both of LSTM-based and transformer-based image captioning framework perform worse than STT series for all seven metrics. It is interesting that LSTM-based model appears to be inferior to transformer-based models, given that transformer frees the model from sequential input dependence. Furthermore, all of modules from STT series tend to improve the results than traditional methods. Besides,  MS (Tensor Mixture Synthesizer) seems to achieve most of the best results among the model zoo. 

\begin{table}[H]
\caption{Image captioning performance with Scaled-Dot-Product self-attention being replaced by STT series.\label{tab:imagecapresult}}
\begin{adjustwidth}{-\extralength}{0cm}
\newcolumntype{c}{>{\centering\arraybackslash}X}
\begin{tabularx}{\fulllength}{C|CCCCCCC}
    \toprule
    \textbf{Meric}   &\textbf{BLEU-1}   &\textbf{BLEU-2}   &\textbf{BLEU-3}   &\textbf{BLEU-4}   &\textbf{METEOR}   &\textbf{ROUGE}   &\textbf{CIDEr}\\
    \midrule
    LSTM       &0.7191  &0.5153  &0.3468  &0.2339  &0.2725  &0.5699  &0.8128 \\
    Transformer&0.7077  &0.5035  &0.3392  &0.2289  &0.2692  &0.5621  &0.7970 \\
    \hline
    SD         &0.7921  &0.6566  &\textbf{0.5386}  &\textbf{0.4457}  &0.3686  &0.7211  &\textbf{1.2099}\\
    SR         &0.7888  &0.6542  &0.5359  &0.4433  &0.3688  &\textbf{0.7213}  &1.2040\\
    MS         &\textbf{0.7925}  &\textbf{0.6569}  &0.5383  &0.4451  &\textbf{0.3691}  &0.7225  &0.2104\\
    FSR        &0.7876  &0.6505  &0.5293  &0.4345  &0.3646  &0.7159  &1.1793\\
    \bottomrule
\end{tabularx}
\end{adjustwidth}
\end{table}

Moreover, in comparison with baseline caption models with LSTM and Transformer,  both SR (Tensor Random Synthesizer) and FSR (Tensor Factorized Random Synthesizer) yield better results with less time for testing, and fewer parameters.
The alignments in transformer are conducted on one side of the feature map while dense synthesizer enables a competitive attention for three sides of the feature tensor, which might result in the overall performance superiority of STT against the baselines.

\section{Discussion}

\subsection{Fewer Computation and Better Extensibility}

In this paper, we present a self-attention plug-in module with its variants, named the series of STT. It employs the tensor transformation for self-alignment to replace the traditional dot-product multiplication in the self-attention module. 
Through this novel approach, the shortage of exhaustive and redundant computation caused by dot-production is dramatically mitigated. 
In our image caption experiments, the STT series achieve most of the best results with less time in testing, and use fewer parameters. 
Moreover, because the synthesizing mechanism notably reduces the input feature's reliance, the underlying structure is preserved. 
The STT series strongly facilitate the robustness of the baseline model. 
In the image classification experiments, the STT series represent the competitive robustness encountering multiple external perturbations. 
It demonstrates the STT series have the potential to improve the robustness, lower the overfitting, and the flexibility to be extended to more deep learning architectures. 

\subsection{Shortage and Future Works}

The STT series improve the efficiency of self-attention module While keeping a competitive performance at the same time, which shows the robustness of the proposed method. However, the performance of this method could be impaired under significant dimension reduction, which should be avoided. Information loss is also inevitably introduced in the process of parameter matrix decomposition. Future works will investigate deeper into the influence of dimension reduction and parameter decomposition on the performance of models and explore possible alternatives for parameter decomposition.

\authorcontributions{Conceptualization, G.Z. and J.Z.; methodology, H.L.and G.Z.; software, G.Z. and Y.F.; validation, J.Z. and Y.F; writing---original draft preparation, H.L.; writing---review and editing, G.Z.; visualization, G.Z.; supervision, J.Z.; project administration, J.Z.; funding acquisition, J.Z. All authors have read and agreed to the published version of the manuscript.}

\funding{This research was funded by General Program of Shanghai Natural Science Foundation (Grant No. 23ZR1419300), Science and Technology Commission of Shanghai Municipality (Grant No. 22DZ2229004)}

\institutionalreview{Not applicable}

\informedconsent{Not applicable}

\dataavailability{No new data were created or analyzed in this study. Data sharing is not applicable to this article}


\conflictsofinterest{The authors declare no conflicts of interest.} 

\begin{adjustwidth}{-\extralength}{0cm}

\reftitle{References}
\externalbibliography{yes}
\bibliography{Definitions/egbib}

\begin{thebibliography}{999}

\bibitem[Roy et~al.(2021)Roy, Saffar, Vaswani, and Grangier]{routeformer}
Roy, A.; Saffar, M.; Vaswani, A.; Grangier, D.
\newblock Efficient Content-Based Sparse Attention with Routing Transformers.
\newblock {\em Trans. Assoc. Comput. Linguistics} {\bf 2021}, {\em 9},~53--68.

\bibitem[Zaheer et~al.(2020)Zaheer, Guruganesh, Dubey, Ainslie, Alberti, Ontanon, Pham, Ravula, Wang, Yang, and Ahmed]{bigbird}
Zaheer, M.; Guruganesh, G.; Dubey, A.; Ainslie, J.; Alberti, C.; Ontanon, S.; Pham, P.; Ravula, A.; Wang, Q.; Yang, L.;  et~al.
\newblock Big Bird: Transformers for Longer Sequences.
\newblock In Proceedings of the NeurIPS,  2020.

\bibitem[Kitaev et~al.(2020)Kitaev, Kaiser, and Levskaya]{reformer}
Kitaev, N.; Kaiser, L.; Levskaya, A.
\newblock Reformer: The Efficient Transformer.
\newblock In Proceedings of the ICLR,  2020.

\bibitem[Wu et~al.(2020)Wu, Lan, Gu, and Yu]{memformer}
Wu, Q.; Lan, Z.; Gu, J.; Yu, Z.
\newblock Memformer: The Memory-Augmented Transformer,  2020,  \href{http://arxiv.org/abs/2010.06891}{{\normalfont [arXiv:cs.CL/2010.06891]}}.

\bibitem[Choromanski et~al.(2021)Choromanski, Likhosherstov, Dohan, Song, Gane, Sarl{\'{o}}s, Hawkins, Davis, Mohiuddin, Kaiser, Belanger, Colwell, and Weller]{performer}
Choromanski, K.M.; Likhosherstov, V.; Dohan, D.; Song, X.; Gane, A.; Sarl{\'{o}}s, T.; Hawkins, P.; Davis, J.Q.; Mohiuddin, A.; Kaiser, L.;  et~al.
\newblock Rethinking Attention with Performers.
\newblock In Proceedings of the ICLR,  2021.

\bibitem[Rae et~al.(2020)Rae, Potapenko, Jayakumar, Hillier, and Lillicrap]{compressive}
Rae, J.W.; Potapenko, A.; Jayakumar, S.M.; Hillier, C.; Lillicrap, T.P.
\newblock Compressive Transformers for Long-Range Sequence Modelling.
\newblock In Proceedings of the ICLR,  2020.

\bibitem[Zhu et~al.(2021)Zhu, Su, Lu, Li, Wang, and Dai]{detectdeformable}
Zhu, X.; Su, W.; Lu, L.; Li, B.; Wang, X.; Dai, J.
\newblock Deformable {DETR:} Deformable Transformers for End-to-End Object Detection.
\newblock In Proceedings of the ICLR,  2021.

\bibitem[Huang et~al.(2019)Huang, Wang, Chen, and Wei]{huang2019attention_AOA_iccv}
Huang, L.; Wang, W.; Chen, J.; Wei, X.
\newblock Attention on Attention for Image Captioning.
\newblock In Proceedings of the ICCV,  2019, pp. 4633--4642.

\bibitem[Fahim et~al.(2020)Fahim, Sarker, Sarker, Sheikh, and Das]{FAHIM2020106437}
Fahim, S.R.; Sarker, Y.; Sarker, S.K.; Sheikh, M.R.I.; Das, S.K.
\newblock Self attention convolutional neural network with time series imaging based feature extraction for transmission line fault detection and classification.
\newblock {\em Electric Power Systems Research} {\bf 2020}, {\em 187},~106437.
\newblock {\url{https://doi.org/https://doi.org/10.1016/j.epsr.2020.106437}}.

\bibitem[Tong et~al.(2024)Tong, Guan, Zhang, Li, Ma, Wu, and Zhu]{tong2024edge}
Tong, W.; Guan, X.; Zhang, M.; Li, P.; Ma, J.; Wu, E.Q.; Zhu, L.M.
\newblock Edge-assisted epipolar transformer for industrial scene reconstruction.
\newblock {\em IEEE Transactions on Automation Science and Engineering} {\bf 2024}.

\bibitem[Mnih et~al.(2014)Mnih, Heess, Graves, et~al.]{rvattention}
Mnih, V.; Heess, N.; Graves, A.;  et~al.
\newblock Recurrent models of visual attention.
\newblock In Proceedings of the NeurIPS,  2014, pp. 2204--2212.

\bibitem[Ba et~al.(2015)Ba, Mnih, and Kavukcuoglu]{Ba2015MultipleOR}
Ba, J.; Mnih, V.; Kavukcuoglu, K.
\newblock Multiple Object Recognition with Visual Attention.
\newblock In Proceedings of the ICLR,  2015.

\bibitem[Wang et~al.(2017)Wang, Jiang, Qian, Yang, Li, Zhang, Wang, and Tang]{residualatt}
Wang, F.; Jiang, M.; Qian, C.; Yang, S.; Li, C.; Zhang, H.; Wang, X.; Tang, X.
\newblock Residual Attention Network for Image Classification.
\newblock In Proceedings of the CVPR,  2017.

\bibitem[Hu et~al.(2018)Hu, Shen, and Sun]{SEnet}
Hu, J.; Shen, L.; Sun, G.
\newblock Squeeze-and-Excitation Networks.
\newblock In Proceedings of the CVPR,  2018, pp. 7132--7141.

\bibitem[Woo et~al.(2018)Woo, Park, Lee, and Kweon]{cbam}
Woo, S.; Park, J.; Lee, J.Y.; Kweon, I.S.
\newblock CBAM: Convolutional Block Attention Module.
\newblock In Proceedings of the ECCV,  2018.

\bibitem[Tay et~al.(2020)Tay, Bahri, Metzler, Juan, Zhao, and Zheng]{synthesizer}
Tay, Y.; Bahri, D.; Metzler, D.; Juan, D.C.; Zhao, Z.; Zheng, C.
\newblock Synthesizer: Rethinking Self-Attention in Transformer Models,  2020,  \href{http://arxiv.org/abs/2005.00743}{{\normalfont [arXiv:cs.CL/2005.00743]}}.

\bibitem[Dong et~al.(2021)Dong, Cordonnier, and Loukas]{not2021attention}
Dong, Y.; Cordonnier, J.; Loukas, A.
\newblock Attention is not all you need: pure attention loses rank doubly exponentially with depth.
\newblock In Proceedings of the ICML,  2021, Vol. 139, pp. 2793--2803.

\bibitem[Liu et~al.(2017)Liu, Mao, Sha, and Yuille]{attcorrection}
Liu, C.; Mao, J.; Sha, F.; Yuille, A.L.
\newblock Attention Correctness in Neural Image Captioning.
\newblock In Proceedings of the AAAI,  2017, pp. 4176--4182.

\bibitem[Wang et~al.(2024)Wang, Ma, and Guo]{app14177794}
Wang, Y.; Ma, N.; Guo, Z.
\newblock Machine Reading Comprehension Model Based on Fusion of Mixed Attention.
\newblock {\em Applied Sciences} {\bf 2024}, {\em 14}.
\newblock {\url{https://doi.org/10.3390/app14177794}}.

\bibitem[Hu et~al.(2018)Hu, Shen, Albanie, Sun, and Vedaldi]{gatherexcite}
Hu, J.; Shen, L.; Albanie, S.; Sun, G.; Vedaldi, A.
\newblock Gather-Excite: Exploiting Feature Context in Convolutional Neural Networks.
\newblock In Proceedings of the NeurIPS,  2018, pp. 9423--9433.

\bibitem[Chen et~al.(2017)Chen, Zhang, Xiao, Nie, Shao, Liu, and Chua]{chen2017sca_cvpr}
Chen, L.; Zhang, H.; Xiao, J.; Nie, L.; Shao, J.; Liu, W.; Chua, T.S.
\newblock Sca-cnn: Spatial and channel-wise attention in convolutional networks for image captioning.
\newblock In Proceedings of the CVPR,  2017, pp. 5659--5667.

\bibitem[Fu et~al.(2019)Fu, Liu, Tian, Li, Bao, Fang, and Lu]{dualatt}
Fu, J.; Liu, J.; Tian, H.; Li, Y.; Bao, Y.; Fang, Z.; Lu, H.
\newblock Dual Attention Network for Scene Segmentation.
\newblock In Proceedings of the CVPR,  2019, pp. 3146--3154.

\bibitem[Bello et~al.(2019)Bello, Zoph, Vaswani, Shlens, and Le]{bello2019attention_augment_iccv}
Bello, I.; Zoph, B.; Vaswani, A.; Shlens, J.; Le, Q.V.
\newblock Attention augmented convolutional networks.
\newblock In Proceedings of the ICCV,  2019, pp. 3286--3295.

\bibitem[Parmar et~al.(2019)Parmar, Ramachandran, Vaswani, Bello, Levskaya, and Shlens]{standalone}
Parmar, N.; Ramachandran, P.; Vaswani, A.; Bello, I.; Levskaya, A.; Shlens, J.
\newblock Stand-Alone Self-Attention in Vision Models.
\newblock In Proceedings of the NeurIPS,  2019, pp. 68--80.

\bibitem[Cordonnier et~al.(2020)Cordonnier, Loukas, and Jaggi]{Cordonnier2020On_relation_iclr}
Cordonnier, J.B.; Loukas, A.; Jaggi, M.
\newblock On the Relationship between Self-Attention and Convolutional Layers.
\newblock In Proceedings of the ICLR,  2020.

\bibitem[Zhao et~al.(2020)Zhao, Jia, and Koltun]{zhao2020exploring_CVPR}
Zhao, H.; Jia, J.; Koltun, V.
\newblock Exploring self-attention for image recognition.
\newblock In Proceedings of the CVPR,  2020, pp. 10076--10085.

\bibitem[Maur{\'i}cio et~al.(2023)Maur{\'i}cio, Domingues, and Bernardino]{Maurcio2023ComparingVT}
Maur{\'i}cio, J.; Domingues, I.; Bernardino, J.
\newblock Comparing Vision Transformers and Convolutional Neural Networks for Image Classification: A Literature Review.
\newblock {\em Applied Sciences} {\bf 2023}.

\bibitem[Hu et~al.(2018)Hu, Gu, Zhang, Dai, and Wei]{relationOD}
Hu, H.; Gu, J.; Zhang, Z.; Dai, J.; Wei, Y.
\newblock Relation Networks for Object Detection.
\newblock In Proceedings of the CVPR,  2018, pp. 3588--3597.

\bibitem[Carion et~al.(2020)Carion, Massa, Synnaeve, Usunier, Kirillov, and Zagoruyko]{endtoendOD}
Carion, N.; Massa, F.; Synnaeve, G.; Usunier, N.; Kirillov, A.; Zagoruyko, S.
\newblock End-to-End Object Detection with Transformers.
\newblock In Proceedings of the ECCV,  2020, Vol. 12346, pp. 213--229.

\bibitem[Wang et~al.(2020)Wang, Zhu, Green, Adam, Yuille, and Chen]{axialsegmentation}
Wang, H.; Zhu, Y.; Green, B.; Adam, H.; Yuille, A.; Chen, L.C.
\newblock Axial-DeepLab: Stand-Alone Axial-Attention for Panoptic Segmentation,  2020,  \href{http://arxiv.org/abs/2003.07853}{{\normalfont [arXiv:cs.CV/2003.07853]}}.

\bibitem[Ondeng et~al.(2023)Ondeng, Ouma, and Akuon]{app131911103}
Ondeng, O.; Ouma, H.; Akuon, P.
\newblock A Review of Transformer-Based Approaches for Image Captioning.
\newblock {\em Applied Sciences} {\bf 2023}, {\em 13}.
\newblock {\url{https://doi.org/10.3390/app131911103}}.

\bibitem[Tong et~al.(2024)Tong, Zhang, Zhu, Xu, and Wu]{tong2024robust}
Tong, W.; Zhang, M.; Zhu, G.; Xu, X.; Wu, E.Q.
\newblock Robust Depth Estimation Based on Parallax Attention for Aerial Scene Perception.
\newblock {\em IEEE Transactions on Industrial Informatics} {\bf 2024}.

\bibitem[Parmar et~al.(2018)Parmar, Vaswani, Uszkoreit, Kaiser, Shazeer, Ku, and Tran]{imtransformer}
Parmar, N.; Vaswani, A.; Uszkoreit, J.; Kaiser, L.; Shazeer, N.; Ku, A.; Tran, D.
\newblock Image transformer.
\newblock In Proceedings of the ICML,  2018, pp. 4055--4064.

\bibitem[Ho et~al.(2019)Ho, Kalchbrenner, Weissenborn, and Salimans]{axialtransformer}
Ho, J.; Kalchbrenner, N.; Weissenborn, D.; Salimans, T.
\newblock Axial attention in multidimensional transformers.
\newblock {\em arXiv preprint arXiv:1912.12180} {\bf 2019}.

\bibitem[Chen et~al.(2020)Chen, Radford, Child, Wu, Jun, Luan, and Sutskever]{igpt}
Chen, M.; Radford, A.; Child, R.; Wu, J.; Jun, H.; Luan, D.; Sutskever, I.
\newblock Generative pretraining from pixels.
\newblock In Proceedings of the ICML,  2020, pp. 1691--1703.

\bibitem[Dosovitskiy et~al.(2021)Dosovitskiy, Beyer, Kolesnikov, Weissenborn, Zhai, Unterthiner, Dehghani, Minderer, Heigold, Gelly, Uszkoreit, and Houlsby]{ViT}
Dosovitskiy, A.; Beyer, L.; Kolesnikov, A.; Weissenborn, D.; Zhai, X.; Unterthiner, T.; Dehghani, M.; Minderer, M.; Heigold, G.; Gelly, S.;  et~al.
\newblock An Image is Worth 16x16 Words: Transformers for Image Recognition at Scale.
\newblock In Proceedings of the ICLR,  2021.

\bibitem[Huang et~al.(2019{\natexlab{a}})Huang, Yuan, Guo, Zhang, Chen, and Wang]{ISSA}
Huang, L.; Yuan, Y.; Guo, J.; Zhang, C.; Chen, X.; Wang, J.
\newblock Interlaced sparse self-attention for semantic segmentation.
\newblock {\em arXiv preprint arXiv:1907.12273} {\bf 2019}.

\bibitem[Huang et~al.(2019{\natexlab{b}})Huang, Wang, Huang, Huang, Wei, and Liu]{CCnet}
Huang, Z.; Wang, X.; Huang, L.; Huang, C.; Wei, Y.; Liu, W.
\newblock CCNet: Criss-Cross Attention for Semantic Segmentation.
\newblock In Proceedings of the ICCV,  2019, pp. 603--612.

\bibitem[Chen et~al.(2018)Chen, Kalantidis, Li, Yan, and Feng]{a2nets}
Chen, Y.; Kalantidis, Y.; Li, J.; Yan, S.; Feng, J.
\newblock A{\^{}}2-Nets: Double Attention Networks.
\newblock In Proceedings of the NeurIPS,  2018, pp. 350--359.

\bibitem[Wang et~al.(2020)Wang, Li, Khabsa, Fang, and Ma]{linformer}
Wang, S.; Li, B.Z.; Khabsa, M.; Fang, H.; Ma, H.
\newblock Linformer: Self-Attention with Linear Complexity,  2020,  \href{http://arxiv.org/abs/2006.04768}{{\normalfont [arXiv:cs.LG/2006.04768]}}.

\bibitem[De~Lathauwer et~al.(2000)De~Lathauwer, De~Moor, and Vandewalle]{de2000multilinear_siam}
De~Lathauwer, L.; De~Moor, B.; Vandewalle, J.
\newblock A multilinear singular value decomposition.
\newblock {\em SIAM journal on Matrix Analysis and Applications} {\bf 2000}, {\em 21},~1253--1278.

\bibitem[Van~Loan(2000)]{van2000ubiquitous_jcam}
Van~Loan, C.F.
\newblock The ubiquitous Kronecker product.
\newblock {\em Journal of computational and applied mathematics} {\bf 2000}, {\em 123},~85--100.

\bibitem[Seibert et~al.(2015)Seibert, W{\"o}rmann, Gribonval, and Kleinsteuber]{tensorproduct}
Seibert, M.; W{\"o}rmann, J.; Gribonval, R.; Kleinsteuber, M.
\newblock Learning co-sparse analysis operators with separable structures.
\newblock {\em IEEE Transactions on Signal Processing} {\bf 2015}, {\em 64},~120--130.

\bibitem[Vaswani et~al.(2017)Vaswani, Shazeer, Parmar, Uszkoreit, Jones, Gomez, Kaiser, and Polosukhin]{vaswani2017attention_NeurIPS}
Vaswani, A.; Shazeer, N.; Parmar, N.; Uszkoreit, J.; Jones, L.; Gomez, A.N.; Kaiser, {\L}.; Polosukhin, I.
\newblock Attention is all you need.
\newblock In Proceedings of the NeurIPS,  2017, pp. 5998--6008.

\bibitem[Krizhevsky et~al.(2009)Krizhevsky, Hinton, et~al.]{cifar100}
Krizhevsky, A.; Hinton, G.;  et~al.
\newblock {\em Learning multiple layers of features from tiny images}; Master Thesis,  2009.

\bibitem[Chen et~al.(2015)Chen, Fang, Lin, Vedantam, Gupta, Doll{\'a}r, and Zitnick]{MSCOCO}
Chen, X.; Fang, H.; Lin, T.Y.; Vedantam, R.; Gupta, S.; Doll{\'a}r, P.; Zitnick, C.L.
\newblock Microsoft coco captions: Data collection and evaluation server.
\newblock {\em arXiv preprint arXiv:1504.00325} {\bf 2015}.

\bibitem[Karpathy and Fei-Fei(2015)]{bleu}
Karpathy, A.; Fei-Fei, L.
\newblock Deep visual-semantic alignments for generating image descriptions.
\newblock In Proceedings of the CVPR,  2015, pp. 3128--3137.

\bibitem[Denkowski and Lavie(2014)]{denkowski2014meteor}
Denkowski, M.; Lavie, A.
\newblock Meteor universal: Language specific translation evaluation for any target language.
\newblock In Proceedings of the Proceedings of the ninth workshop on statistical machine translation,  2014, pp. 376--380.

\bibitem[Vedantam et~al.(2015)Vedantam, Lawrence~Zitnick, and Parikh]{vedantam2015cider}
Vedantam, R.; Lawrence~Zitnick, C.; Parikh, D.
\newblock Cider: Consensus-based image description evaluation.
\newblock In Proceedings of the CVPR,  2015, pp. 4566--4575.

\bibitem[Lin(2004)]{lin2004rouge}
Lin, C.Y.
\newblock Rouge: A package for automatic evaluation of summaries.
\newblock In Proceedings of the Text summarization branches out,  2004, pp. 74--81.

\end{thebibliography}




%


\PublishersNote{}
\end{adjustwidth}
\end{document}